# Research on Effectiveness Evaluation and Optimization of Baseball Teaching Method Based on Machine Learning

**Shaoxuan Sun [1,4], Jingao Yuan [2,5], Yuelin Yang[3,6]**

[1] Physical Education Department, Shandong University (Weihai), Shandong Province, China
[2] Department of physical education, Xiamen University, Fujian Province, China
[3] Department of Kinesiology, California Baptist University, California, US

[4] ssxuan@sdu.edu.cn
[5] 26920241153893@stu.xmu.edu.cn
[6] 769006@calbaptist.edu

**Abstract.** In modern physical education, data-driven evaluation methods have gradually attracted attention, especially the quantitative prediction of students' sports performance through machine learning model. The purpose of this study is to use a variety of machine learning models to regress and predict students' comprehensive scores in baseball training, so as to evaluate the effectiveness of the current baseball teaching methods and put forward targeted training optimization suggestions. We set up a model and evaluate the performance of students by collecting many characteristics, such as hitting times, running times and batting. The experimental results show that K-Neighbors Regressor and Gradient Boosting Regressor are excellent in comprehensive prediction accuracy and stability, and the R score and error index are significantly better than other models. In addition, through the analysis of feature importance, it is found that cumulative hits and cumulative runs are the key factors affecting students' comprehensive scores. Based on the results of this study, this paper puts forward some suggestions on optimizing training strategies to help students get better performance in baseball training. The results show that the data-driven teaching evaluation method can effectively support physical education and promote personalized and refined teaching plan design.

## 1. Introduction

In physical education, how to effectively evaluate and improve students' comprehensive performance has always been the focus of research and practice. Traditional evaluation methods often rely on teachers' subjective judgment and simple record of achievements, and it is difficult to comprehensively and objectively reflect students' real ability and growth potential. With the rapid development of data science and artificial intelligence, quantitative analysis of sports performance by using machine learning technology has gradually become a new trend. Machine learning model can not only process multi-dimensional data, but also mine patterns hidden in features to support personalized teaching.

In baseball training, hitting, running and batting are the core elements to measure students' comprehensive performance. In view of these characteristics, this study collected students' performance data in the training process, and used a variety of regression models to predict and analyze, in order to explore the influence of these characteristics on students' comprehensive scores. Specifically, we choose K-Neighbors Regressor, Gradient Boosting Regressor, SVM and other machine learning models to compare their performance in regression prediction, and identify key influencing factors through feature importance analysis. In the experiment, the generalization ability of the model is also evaluated by K-fold cross-validation to ensure the stability and reliability of the prediction results.

This study not only evaluates the effectiveness of baseball training through data-driven methods, but also provides reference for how to optimize teaching methods. The results show that specific training indicators (such as cumulative hits and cumulative runs) have a significant impact on students' comprehensive scores, which provides data basis for further improving training plans. Through this study, we hope to provide a scientific evaluation framework for baseball and other physical education, and promote the development of personalized and data-driven physical education.

2. literature review

In recent years, the application of machine learning technology in sports analysis has attracted wide attention, especially in the field of baseball. Baseball has become an ideal scene for machine learning research because of its rich data and information. This paper refers to many research documents about the application of machine learning in baseball field, and sorts out the related research progress and achievements.

Hamilton et al. discussed the application of machine learning in baseball pitching prediction, aiming at predicting the possibility of pitching by analyzing pitching data and opponents' performance [1]. This study adopts a variety of machine learning algorithms, including support vector machine and random learning, which have potential in pitching prediction and strategy formulation, and provide assistant decision support for coaches and pitchers.

Koseler and Stephan made a systematic literature review on the application of machine learning in baseball. They summarized the main machine learning applications, including player performance prediction, game result prediction, stadium decision support and so on [2]. By processing a large number of historical data, they can find potential trends and laws and help management and coaches optimize their decisions. However, traditional statistical methods still dominate in some applications, so the further development of this field needs to introduce machine learning more systematically.

With the deepening of research, Karnuta et al. demonstrated the advantages of machine learning in predicting players' injuries [3]. They compared machine learning with traditional regression analysis, and the results show that machine learning model can predict the injury risk of professional baseball players more accurately. Based on 13,98 studies, the excellent performance of machine learning model in big data processing and pattern recognition is confirmed, which shows great potential in pre-season planning and player management.

In the prediction of competition results, Huang and Li applied machine learning and deep learning technology to build a model for predicting competition results. Their research found that the deep learning model is complex to traditional methods, especially when the data characteristics are highly nonlinear. By inputting a large number of game data into the neural network model, this study predicts the game results, achieving high accuracy, which shows the broad prospects of deep learning in sports prediction [4].

In addition, Yaseen et al. proposed a multi-modal machine learning method to predict the playoff results of professional baseball leagues [5]. This study combines a variety of data sources, including team performance, player performance, historical game data, etc., and uses prediction accuracy [6].

This study shows the potential application of multi-modal data fusion technology in sports event prediction, which provides a new direction for future research [7].

To sum up, the application of machine learning in baseball shows broad prospects and practical value. Although there are still challenges in data quality and feature engineering, the existing research has proved the advantages of machine learning in game prediction, player performance analysis, injury prediction and so on. In the future, with the development of advanced technologies such as deep learning and multimodal learning, the application of machine learning in baseball will be further expanded, and it will play an important role in improving decision support, improving game strategy and ensuring players' health.

## 3. Data

*3.1 Data introduction*

This dataset contains a variety of features that reflect the performance and progress of students in their baseball training. The data can be divided into several categories:

Current Performance Data: Features such as AtBat, Hits, HmRun, and Runs reflect the student's actual performance in baseball during the current semester. These indicators provide a snapshot of the student's training level and can be used to assess their current abilities, highlighting their strengths and areas for improvement.

Cumulative Data: Features like CAtBat, CHits, CHmRun, and CRuns represent the student's cumulative training data. These values offer insight into their long-term performance over time, allowing for a better understanding of the student's development. By analyzing these cumulative statistics, trends in the student's progress can be identified, and their overall skill level can be assessed.

Defensive Performance: The features PutOuts, Assists, and Errors describe the student's defensive performance during training. PutOuts measures the number of outs the student has made, Assists records the number of assists provided to teammates in achieving outs, and Errors tracks mistakes made in the field. Together, these features offer a comprehensive view of the student's abilities and consistency in defense.

Target Variable: The score feature serves as the target variable, representing the overall performance level of the student. The primary goal of the machine learning model is to predict the score based on the other features, identifying which aspects of the student's training (e.g., batting, defense, experience) most influence their final score. This allows for a data-driven assessment of how well students are performing and provides actionable feedback for improving teaching methods.

Overall, this dataset offers a comprehensive view of student performance in various facets of baseball training, including both offensive and defensive skills. By using machine learning models to analyze these features, we can gain a deeper understanding of how different aspects of training contribute to a student's overall score. This information can be used to refine teaching strategies and provide tailored feedback to help students improve their skills.

**Table 1.** Variable definition

| Feature | Description |
|---|---|
| AtBat | The number of times a player has been at bat during the training session this semester. |
| Hits | The number of successful hits (base hits) during the training session this semester. |
| HmRun | The number of home runs during the training session this semester. |
| Runs | The number of times a player has run the bases and scored during the training session this semester. |
| RBI | The number of runs batted in (RBIs) during the training session this semester. |
| Walks | The number of walks (four balls) a player received during the training session this semester. |
| years | The total number of years a player has participated in baseball training. |
| CAtBat | The cumulative number of times the player has been at bat over their training career. |
| CHits | The cumulative number of successful hits (base hits) over the player's training career. |
| CHmRun | The cumulative number of home runs over the player's training career. |
| CRuns | The cumulative number of runs scored over the player's training career. |
| CRBI | The cumulative number of RBIs over the player's training career. |
| CWalks | The cumulative number of walks (four balls) the player has received during their training career. |
| PutOuts | The number of putouts the player has made during the training session this semester. |
| Assists | The number of assists (helping teammates to make an out) the player has made during the training session this semester. |
| Errors | The number of errors (mistakes) the player has made during the training session this semester. |
| score | The overall performance score, which evaluates the player's overall level based on their training statistics. |

*3.2 Descriptive statistical analysis*

The variable AtBat has a total of 322 observations. The mean value is 380.93, with a standard deviation of 153.41, indicating a moderate spread of values around the mean. The minimum value is 16, and the maximum is 687, suggesting that some students have very low and very high participation in training. The distribution is positively skewed, as shown by the increasing percentiles, with the 50th percentile (median) being 379.50. This suggests that half of the students have fewer than 380 at-bats, while the other half have more. The 99th percentile value is 658.59, which is close to the maximum, further supporting the presence of a few students with significantly higher training participation.

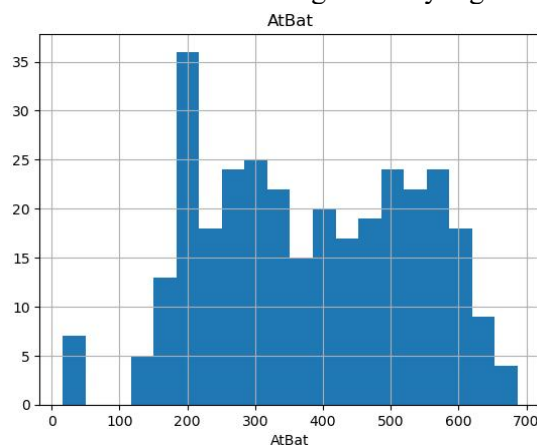

**Figure 1.** Statistical analysis of AtBat variable description

The variable Hits has 322 observations, with a mean value of 101.03 and a standard deviation of 46.46, indicating a moderate variation in the number of successful hits among students. The minimum value is 1, while the maximum value is 238, reflecting a wide range of performance. The distribution is slightly skewed towards higher values, as the percentiles increase gradually. For example, the median value (50th percentile) is 96, with 50% of the students having fewer than 96 hits and the other half having more. The 99th percentile value is 210.79, which is close to the maximum, highlighting a small group of students who perform significantly better than the majority.

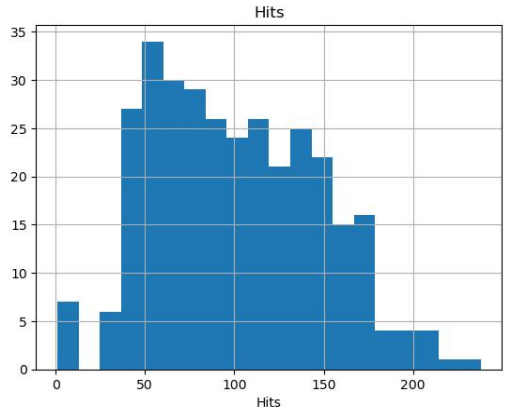

**Figure 2.** Statistical analysis of Hits variable description

## 4. Model analysis

*4.1 model introduction*
Support Vector Machine (SVM): SVM is a supervised learning algorithm commonly used for classification and regression tasks. It works by finding the hyperplane that best separates data points into different classes or predicts continuous outcomes by maximizing the margin between data points and the decision boundary. In regression tasks, SVM seeks to minimize the error within a certain tolerance level, offering robust results in high-dimensional spaces.

K-Neighbors Regressor: This model is based on the k-nearest neighbors (KNN) algorithm, which uses the similarity between data points to make predictions. For regression, it predicts the value of a new data point based on the average values of its k-nearest neighbors in the feature space. The model is straightforward, making it interpretable and effective for certain types of datasets.

Kernel Ridge Regression: This model combines ridge regression with kernel methods, allowing it to model non-linear relationships by mapping input features into a higher-dimensional space. Kernel Ridge Regression applies regularization to prevent overfitting, making it suitable for datasets with complex relationships between features.

Decision Tree Regressor: This model uses a tree-like structure to make decisions and predictions. At each node, the model splits data based on specific criteria to minimize prediction error. Decision trees are intuitive and easy to interpret, though they can be prone to overfitting if not properly regularized.

Random Forest Regressor: Random Forest is an ensemble method that builds multiple decision trees on different subsets of the data and averages their predictions. This approach reduces overfitting and improves generalization. It's widely used due to its high accuracy and robustness on large datasets [8].

Logistic Regression: Although primarily used for classification, logistic regression can also be adapted for regression tasks. It models the relationship between features and a target variable by estimating probabilities using a logistic function. Its simplicity and interpretability make it a common baseline model in machine learning [9].

Gradient Boosting Regressor: This model is a boosting technique that builds an ensemble of weak learners (typically decision trees) sequentially. Each tree corrects the errors of the previous one, gradually improving prediction accuracy. Gradient Boosting is powerful for handling complex, non-linear relationships in data [10].

*4.2 Model results*

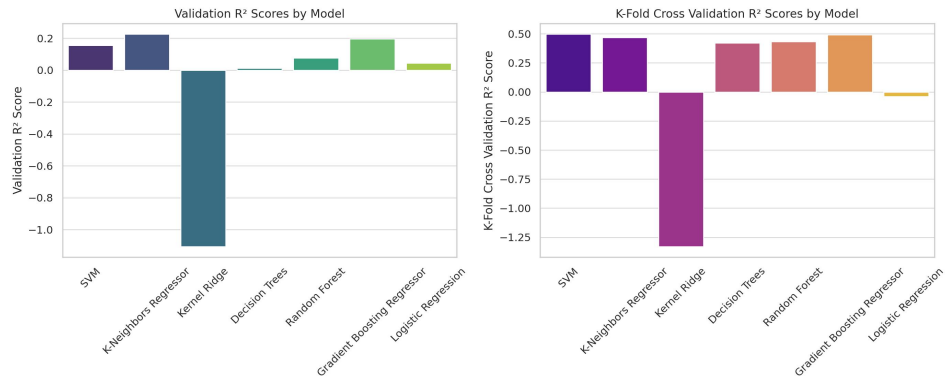

**Figure 3.** Validation R-squared results of each model

K-Neighbors Regressor and Gradient Boosting Regressor: These two models perform relatively well. On the verification set, K-Neighbors Regressor got 22.7% R score, and MAE and RMSE were 254.44 and 373.92 respectively. The verification R score of GradientBoosting Regression is 19.75%, and its MAE and RMSE are relatively low, indicating that the prediction of the target variable is more accurate. In addition, the Gradient Boosting Regressor scored high (49.20%) in K-fold cross-validation, indicating that it has good stability under different data segmentation.

The verification R score of SVM and Random Forest：SVM is 15.6%, while Random Forest is 7.67%. Although the performance of SVM is slightly better than Random Forest, their MAE and RMSE are still higher. SVM obtains 49.81% R in K-fold cross-validation, while Random Forest is 43.3%, which indicates that the stability of SVM is slightly higher than Random Forest, but the accuracy of the model in the validation set still needs to be improved.

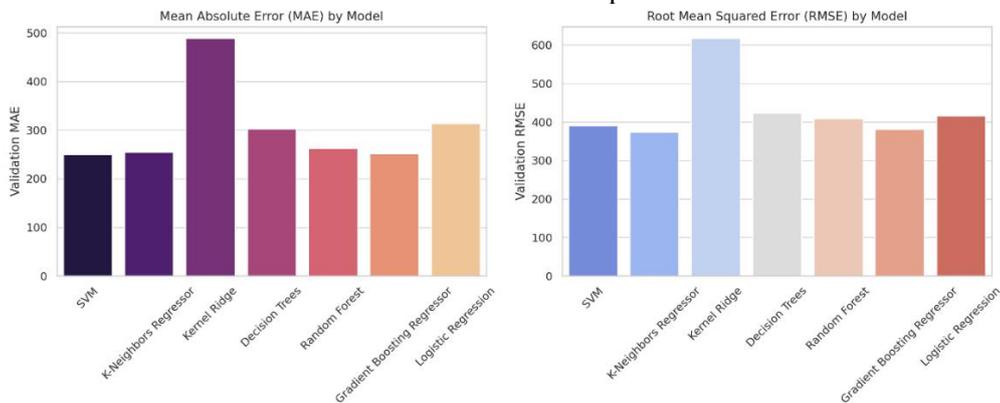

**Figure 4.** MAE and RMSE results of each model

Kernel Ridge and Decision Trees：Kernel Ridge performed poorly, the validation r was -110.59%, and the K-fold cross-validation score was also negative, which reflected their poor fitting effect on this data set. The validation r of Decision Trees is 1.32%, which is also not good, and the standard deviation in validation set and K-fold cross validation is large, indicating that its stability is insufficient.

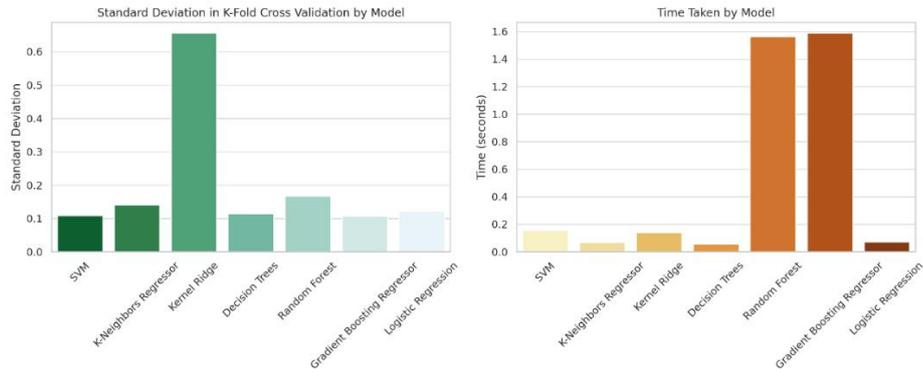

**Figure 5.** Standard deviation and time taken results of each model

Logistic Regression: Although Logistic Regression is usually used in classification tasks, it is mediocre in regression tasks, with a test R score of only 4.42% and a negative score in K-fold cross-test. Logistic Regression is not suitable for regression tasks, so its results are not good.

Generally speaking, K-Neighbors Regressor and Gradient Boosting Regressor have obtained higher R scores and lower MAE and RMSE on the verification set, which are the more recommended models in the current results.

*4.3 Feature importance analysis*

According to the provided feature importance data, we can see the influence of each feature on the prediction results of the model. The following is an interpretation of each feature:

CHits (cumulative hits, importance 0.072): This is the index with the highest feature importance, indicating that cumulative hits play a key role in predicting students' baseball performance. Higher cumulative hits usually indicate students' batting skills and stable batting ability, which is closely related to the overall performance. Because its continuous performance in training and competition has obvious contribution to the evaluation results, it has the highest weight in the model.

CRuns (Cumulative Run Times, Importance 0.07): Followed by Cumulative Run Times, which is slightly less important in the model than CHits. This shows that running ability is also an important factor affecting performance. Cumulative running times directly reflect students' speed, judgment and performance on the court, which is usually the basis for further scoring after hitting the ball successfully, so it has a positive effect on the overall performance of baseball.

RBI (hit points, importance 0.058): hit points measure students' contribution to the team score, although its importance is not as good as the cumulative indicators (CHits and CRuns), it is still significant. This feature reflects the ability of students to hit high-quality balls at critical moments, which often has a decisive impact on the outcome. Getting a higher RBI in training and competition usually means that students can improve their scoring efficiency at critical moments, which has positive guiding significance for the evaluation of teaching methods.

CRBI (Cumulative Hit Points, Importance 0.056): Cumulative Hit Points further show students' contribution in the long-term competition, especially how they let their teammates score by hitting the ball. Although its importance is slightly lower than RBI, it still plays a key role. CRBI reflects students' ability to play their scoring potential for a long time on the field, thus supporting students' comprehensive performance.

CAtBat (cumulative number of hits, importance 0.053): This is the lowest index of feature importance, but it still has some influence. Cumulative batting times not only indicates the number of times students play, but also represents their participation and proficiency in the game. Although its influence is relatively small compared with other features, the increase of CAtBat value can usually indirectly enhance the role of other features (such as CHits and RBI) in the overall performance.

From the analysis of the importance of overall features, the cumulative features (CHits, CRuns, CRBI, CAtBat) have a high weight in the prediction model, which indicates that students' long-term performance accumulation plays a significant role in evaluating their baseball training effect and comprehensive performance. By paying attention to accumulated achievements, the model captures more stable performance characteristics, which are helpful to better evaluate and optimize teaching methods. In addition, the high importance of RBI and CRBI shows that in the improvement of teaching methods, we can pay attention to improving students' scoring ability at critical moments to enhance the positive impact on the overall performance.

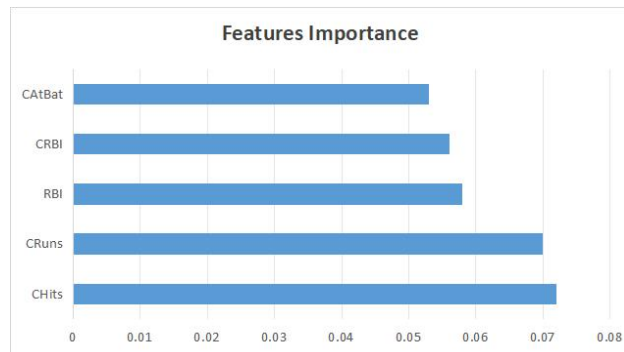

**Figure 6.** Feature importance analysis

**5. Conclusions and Suggestions**

*5.1 Conclusions*
Based on a variety of machine learning models, this study makes a regression prediction of students' comprehensive scores in baseball training, and evaluates the effectiveness of current baseball training methods through model performance and feature importance analysis. The results of multi-model experiments show that K-Neighbors Regression, Gradient Boosting Regressor and SVM perform well in comprehensive prediction performance, and obtain higher verification R score and lower error index, while the prediction accuracy of Logistic Regression and Kernel Ridge models is relatively low. In particular, the Gradient Boosting Regressor has shown good performance in the overall stability and prediction accuracy of the model, indicating its advantages in dealing with multi-feature prediction tasks.

In the analysis of feature importance, features such as cumulative hits (CHits) and cumulative runs (CRuns) have higher weights, which shows that students' long-term performance accumulation and key scoring ability have an important influence on the comprehensive score. This discovery reveals the importance of cultivating students' hitting and running ability to improve their overall performance, and also verifies the rationality of feature selection in model prediction.

In addition, the generalization ability of K-fold cross-validation evaluation model is introduced into machine learning prediction. The results show that the performance of each model has little difference between training data and validation data, indicating that the model has good stability under different data partitions. Through experiments, we also found that proper feature engineering and data preprocessing (such as feature coding, standardization and dimensionality reduction) can improve the performance of the model and help to better capture the key patterns in the training data.

The conclusion of this study not only helps to reveal the key factors that affect students' comprehensive performance in baseball training, but also provides data support for the optimization of baseball teaching methods in the future. Combined with the prediction results of the model and the analysis of feature importance, teachers can improve the training methods in a targeted manner, and improve students' comprehensive scores and overall performance level.

*5.2 Suggestions*
Improve the accuracy of hitting: According to the analysis of characteristic importance, the cumulative number of hits has the greatest influence on students' comprehensive scores. Therefore, students' hitting skills can be emphasized in training, and the success rate and accuracy of hitting can be improved through targeted training such as hitting timing and hand-eye coordination, so as to increase the chances of scoring.

Enhance running ability: running is the key link in baseball game, and the cumulative number of runs ranks second in the feature importance. Therefore, it is suggested to pay attention to the improvement of students' speed, reaction ability and running route judgment in training. This will help students more effectively change from hitting the ball to scoring opportunities, and further improve team performance.

Strengthening the ability to score points: the importance of scoring points and cumulative scoring points in the model is high, which shows that the ability of students to score points at critical moments has an important impact on their overall performance. By simulating the key scenes in the competition, we can cultivate students' ability to hit the ball stably under pressure and let their teammates score points, and help students make correct judgments in the competition.

Increasing the actual combat opportunities: Although the cumulative batting times are slightly lower in the characteristic importance, they still contribute to the prediction of students' comprehensive performance. We can appropriately increase students' playing times, give them more opportunities for actual combat, and let them accumulate experience and improve their ability to adapt to the competition. This suggestion is especially suitable for students who have just started baseball training to help them hone their skills and improve their performance in actual games.

Personalized feedback and progress tracking: Combining the predicted results of the model, provide targeted feedback and training programs to help students understand their own strengths and weaknesses. For example, by analyzing the feature weights, students can be explained in detail their performance in hitting the ball, running and so on, and set clear progress goals. This kind of personalized guidance can help students improve their overall performance more effectively in training.